\documentclass[journal]{IEEEtran}
\usepackage[letterpaper, left=54pt, right=54pt, bottom=54pt, top = 54pt]{geometry}
\newgeometry{left=54pt, right=54pt, bottom=54pt, top = 72pt}
\IEEEoverridecommandlockouts

\usepackage{amsmath}

\usepackage{amsfonts}
\usepackage{hyperref}
\usepackage{textcomp}
\usepackage{subcaption}
\usepackage{xfrac}

\usepackage{tikz}
\usepackage{xfrac}
\usepackage{siunitx}
\usepackage{float}
\usepackage{pgfplots}
\usepackage{tikz-3dplot}
\usetikzlibrary{calc}
\pgfplotsset{width=7cm,compat=1.8}

\usepackage{color}
\usepackage{soul}

\usepackage[style=ieee, citestyle=numeric-comp,sorting=none]{biblatex}
\def\BibTeX{{\rm B\kern-.05em{\sc i\kern-.025em b}\kern-.08em
    T\kern-.1667em\lower.7ex\hbox{E}\kern-.125emX}}

\begin{document}
\newcommand{\braces}[1]{\left\{ #1 \right\}}
\newcommand{\leftbrace}[1]{\left\{ #1 \right.}
\newcommand{\brackets}[1]{\left[ #1 \right]}
\newcommand{\paren}[1]{\left( #1 \right)}
\newcommand{\evalat}[1]{\left. #1 \right|}
\newcommand{\abs}[1]{\left| #1 \right|}
\newcommand{\norm}[1]{\left\|#1\right\|}
\newcommand{\mat}[1]{\ensuremath{\left[#1\right]}}
\newcommand{\bs}[1]{\ensuremath{\boldsymbol{#1}}}
\newcommand{\matb}[1]{\ensuremath{\mat{\bs{#1}}}}

\newcommand{\RNum}[1]{\textbf{\uppercase\expandafter{\romannumeral #1\relax}}}
\newcommand{\one}{\RNum{1}}
\newcommand{\two}{\RNum{2}}
\newcommand{\bfJ}{\mathbf{J}}
\newcommand{\bfF}{\mathbf{F}}
\newcommand{\bfG}{\mathbf{G}}
\newcommand{\Rp}{{\mathbf{R}^{gc}}}
\newcommand{\Rb}{{\mathbf{R}^{gb}}}
\newcommand{\Rr}{{\mathbf{R}^{cb}}}

\newcommand{\Fg}{\bs{\mathcal{F}}^g}
\newcommand{\Fb}{\bs{\mathcal{F}}^b}
\newcommand{\Fp}{\bs{\mathcal{F}}^c}

\newcommand{\xg}{\ensuremath{\bs{x}^g}}
\newcommand{\eb}[1]{\ensuremath{\bs{e}^b_{#1}}}
\newcommand{\eg}[1]{\ensuremath{\bs{e}^g_{#1}}}
\newcommand{\et}[1]{\ensuremath{\bs{e}^t_{#1}}}
\newcommand{\ef}[1]{\ensuremath{\bs{e}^f_{#1}}}
\newcommand{\er}[1]{\ensuremath{\bs{e}^r_{#1}}}
\newcommand{\emot}[1]{\ensuremath{\bs{e}^m_{#1}}}
\newcommand{\esc}{\ensuremath{\bs{e}^c_s}}
\newcommand{\eyc}{\ensuremath{\bs{e}^c_y}}
\newcommand{\enc}{\ensuremath{\bs{e}^c_n}}

\newcommand{\xp}{\ensuremath{\bs{x}^p}}
\newcommand{\xps}{\ensuremath{\bs{x}^p_s}}
\newcommand{\xpsmag}{\norm{\xps}}
\newcommand{\eps}{\ensuremath{\bs{e}^p_s}}
\newcommand{\xpy}{\ensuremath{\bs{x}^p_y}}
\newcommand{\xpymag}{\norm{\xpy}}
\newcommand{\epp}{\ensuremath{\bs{e}^p_\perp}}
\newcommand{\xpn}{\ensuremath{\bs{e}^p_n}}
\newcommand{\epn}{\ensuremath{\bs{e}^p_n}}
\newcommand{\xpss}{\ensuremath{\bs{x}^p_{ss}}}
\newcommand{\xpsy}{\ensuremath{\bs{x}^p_{sy}}}
\newcommand{\xpys}{\ensuremath{\bs{x}^p_{ys}}}
\newcommand{\xpyy}{\ensuremath{\bs{x}^p_{yy}}}

\newcommand{\rcom}{\bs{r}_{\text{com}}}
\newcommand{\vcom}{\bs{v}_{\text{com}}}

\newcommand{\ths}{\theta^s}
\newcommand{\thp}{\theta^p}
\newcommand{\ks}{\kappa^c_s}
\newcommand{\ky}{\kappa^c_y}
\newcommand{\kn}{\kappa^c_n}

\newcommand{\vb}[1]{\ensuremath{v^b_{#1}}}
\newcommand{\vt}[1]{\ensuremath{v^t_{#1}}}
\newcommand{\vf}[1]{\ensuremath{v^f_{#1}}}
\newcommand{\wb}[1]{\ensuremath{\omega^b_{#1}}}
\newcommand{\wm}[1]{\ensuremath{\omega^m_{#1}}}

\title{A General 3D Road Model for Motorcycle Racing\\
\thanks{Thomas Fork and Francesco Borrelli are with the Department of Mechanical Engineering, University of California, Berkeley, USA.}
\thanks{Correspondence may be addressed to fork@berkeley.edu.}\thanks{Supported by Brembo Inspiration Lab}
}

\author{\IEEEauthorblockN{Thomas Fork}
and
\IEEEauthorblockN{Francesco Borrelli}
}

\maketitle
\begin{abstract}
We present a novel control-oriented motorcycle model and use it for computing racing lines on a nonplanar racetrack. The proposed model combines recent advances in nonplanar road models with the dynamics of motorcycles. Our approach considers the additional camber degree of freedom of the motorcycle body with a simplified model of the rider and front steering fork bodies. We demonstrate the effectiveness of our model by computing minimum-time racing trajectories on a nonplanar racetrack. 
\end{abstract}


\section{Introduction}

Control-oriented vehicle models have seen widespread use for trajectory planning in consumer \cite{7490340, 7225830} and motorsport \cite{christ2021time, 3d_part_2} applications. However, many such models have been limited to simple road geometry, such as flat roads, roads with constant slope {\cite{rajamani_book}}, straight motion on a banked road {\cite{guiggiani_book}} or simple crests and dips {\cite{reza_book}}. This does not adequately capture vehicle behavior for safety-critical or high performance maneuvers, ubiquitous in both consumer and motorsport industries. These limitations primarily result from the lack of suitable road models \cite{limebeer2023review}. 

Early literature \cite{LOT20147559, 3d_part_1} developed 3D road models for ribbon-shaped surfaces, which may curve and twist in 3D but are cross-sectionally linear. These works focused on four-wheeled vehicles, not motorcycles in part due to their more complicated dynamics and ability to camber. Later work \cite{leonelli2020optimal} applied these road models to motorcycles. However, their work was limited to the same road models, whereas not all roads have flat cross-section and some environments may not have roads at all. In \cite{fork2021models}, the authors developed a general 3D road model involving a near-arbitrary parametric surface. In this paper we apply our road model of \cite{fork2021models} to a simple motorcycle model and use it to compute racelines: periodic minimum-time trajectories around a 3D racetrack.

The remainder of this paper is structured as follows: We introduce background on motorcycles and the proposed road model in Section \ref{sec:background}. We introduce our road model in full in Section \ref{sec:road_model}, with adaptations to the motion of motorcycles. We introduce motorcycle dynamics and our motorcycle model in Section \ref{sec:vehicle_model}. We use our model to compute racelines in Section \ref{sec:control} with results and conclusions in Sections \ref{sec:results} and \ref{sec:conclusion}.

\section{Background} \label{sec:background}

\subsection{Motorcycle Geometry} \label{sec:motorcycle_background}

\begin{figure}
    \centering
    \includegraphics[width=\linewidth]{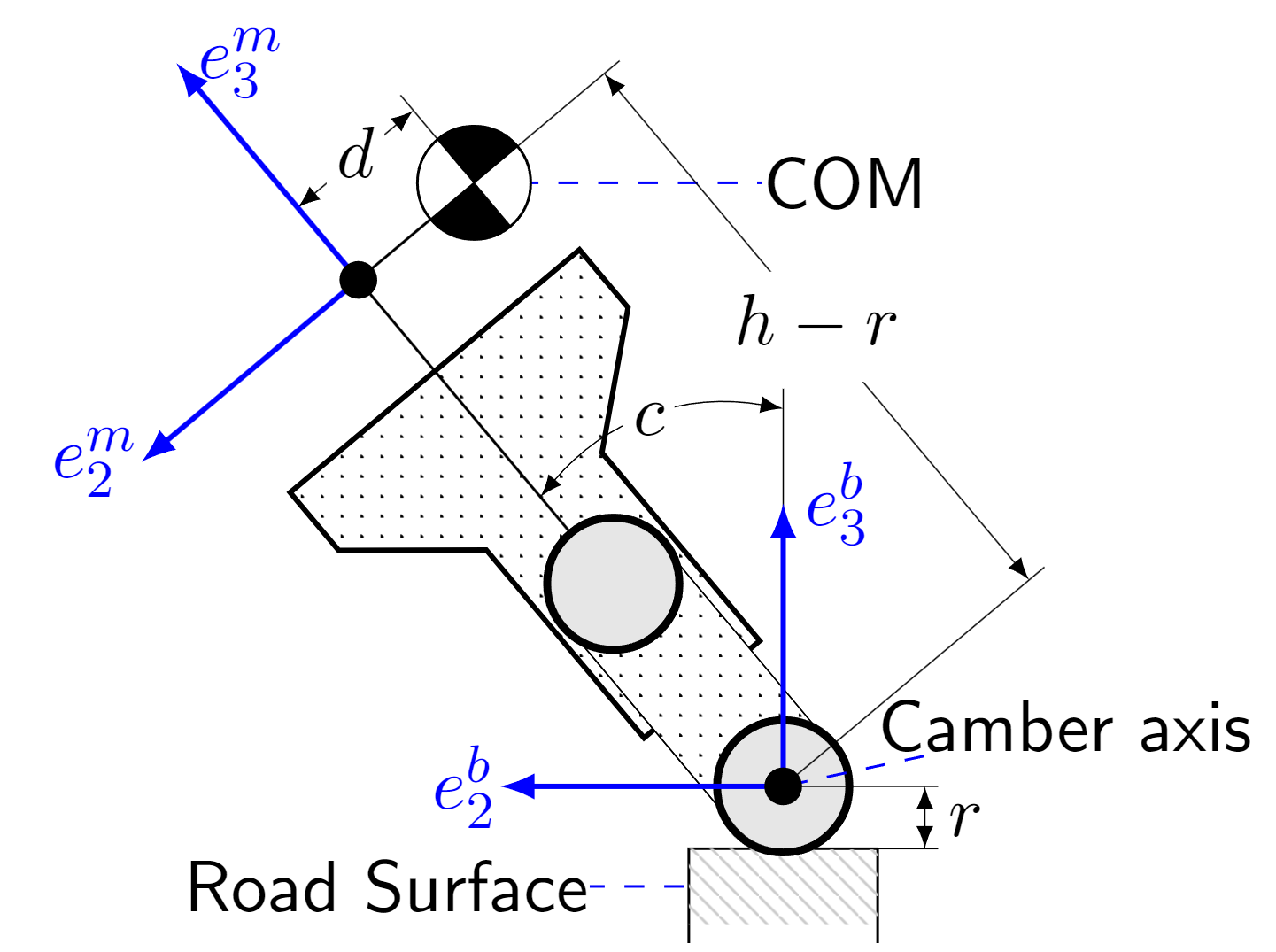}
    \caption{Schematic of motorcycle geometry as viewed from behind. The motorcycle body is assumed to camber with angle $c$ about a point located a distance $r$ above the road surface. The center of mass (COM) is located at height $h-r$ and lateral offset $d$ in the frame of the motorcycle due to motion of the rider's body. $\eb{1} = \emot{1}$ and point into the page. We define $c>0$ and $d>0$ when they shift to the left from the driver's perspective. ($c>0$ and $d<0$ as shown)
    }
    \label{fig:motorcycle_dims}
\end{figure}

Motorcycles are inherently multi-body systems comprised of wheels, front and rear suspension, rider, chassis, and more \cite{tanelli2014modelling}. Control-oriented models invariably simplify some components to capture controllable behaviour while omitting finer details. In the present paper, we make the assumption that there exists an axis fixed relative to the chassis of the motorcycle which remains a constant distance above the road. Our ``camber axis" (Fig. \ref{fig:motorcycle_dims}) implies that the motorcycle does not pitch forwards or backwards, and imposes further assumptions on tires, suspension, and steering geometry.

\begin{figure}
    \centering
    \includegraphics[width=\linewidth]{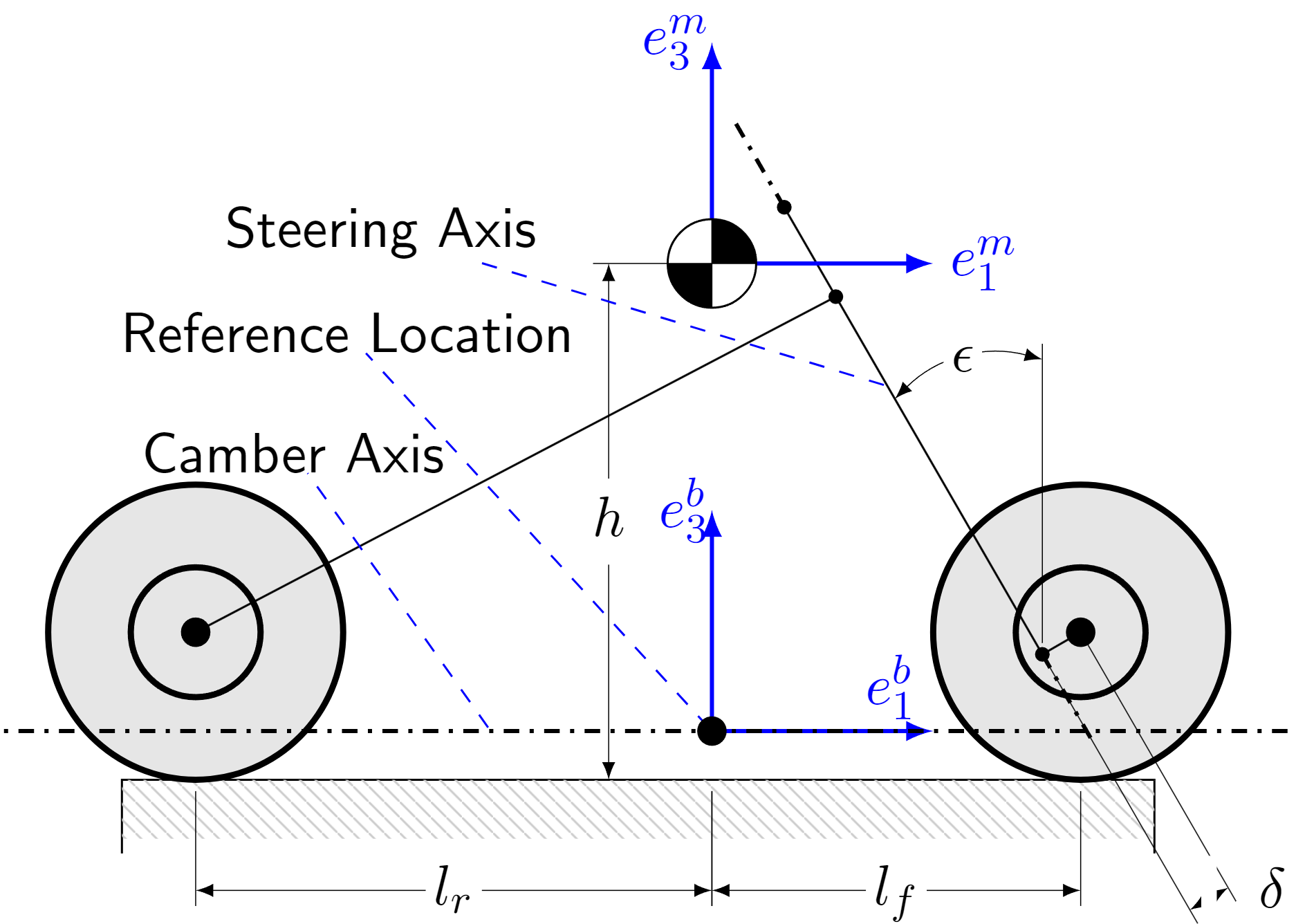}
    \caption{Schematic of motorcycle geometry as viewed from side with no camber. The front steering assembly has rake angle $\epsilon$ and offset $\delta$. The center of mass is a distance $h$ above the road surface, with the reference location directly below it. The front and rear wheels make contact with the surface at distances $l_f$ and $l_r$ along the motorcycle relative to the reference location. $\eb{1}$ and $\emot{1}$ are always equal while $\eb{3}$ and $\emot{3}$ are only equal at zero camber (see Figure \ref{fig:motorcycle_dims}).}
    \label{fig:motorcycle_side_dims}
\end{figure}

The camber axis will be vital to precisely link nonplanar road surface and motorcycle geometry. We use it to introduce a reference location along the camber axis and directly below the center of mass (COM), shown in Figure \ref{fig:motorcycle_side_dims}. We equip our reference location with the orthonormal basis $\eb{1,2,3}$ and refer to the two together as the body frame. Similarly, we introduce a motorcycle frame fixed to the motorcycle chassis at the height of the COM with basis $\emot{1,2,3}$. We allow the COM itself to move laterally in the motorcycle frame due to rider motion as shown in Figure \ref{fig:motorcycle_dims}. 

\begin{figure}
    \centering
    \includegraphics[width=\linewidth]{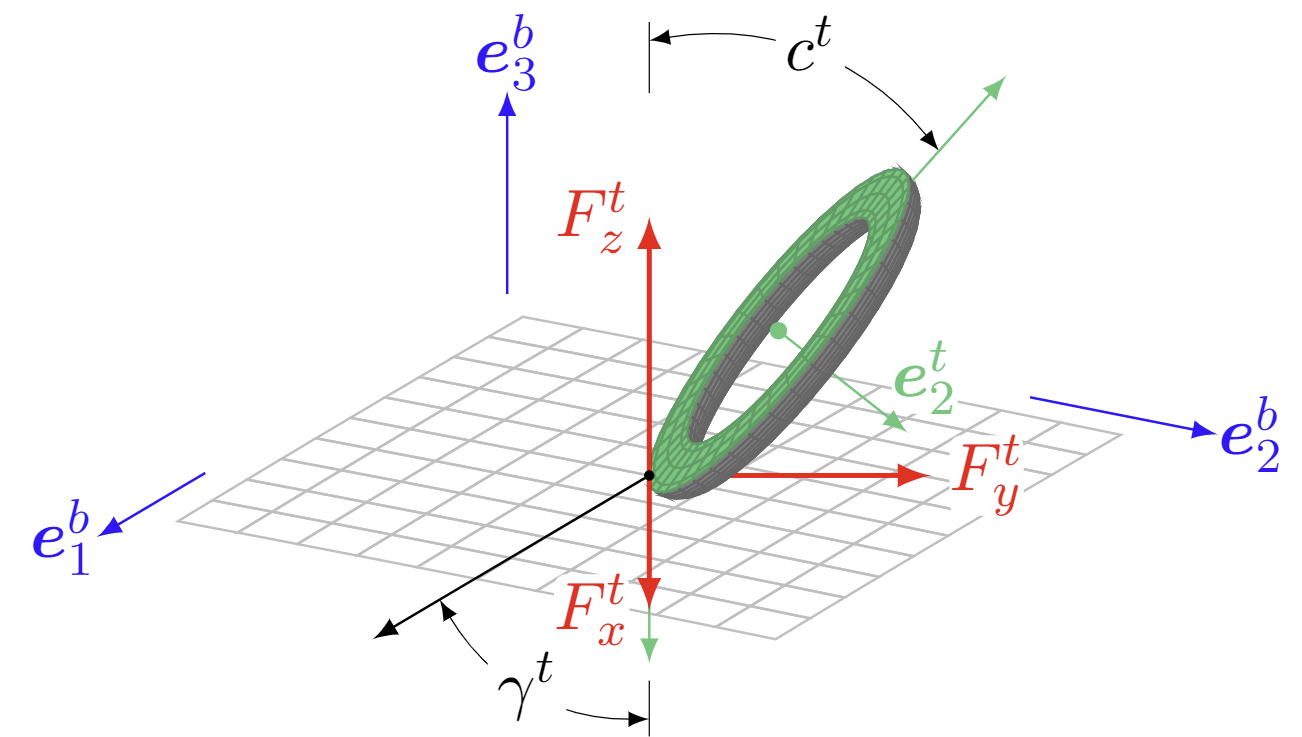}
    \caption{Tire diagram, with the tire cross-sectioned through its plane of symmetry. Tire camber and steering angle $c^t$ and $\gamma^t$ are positive as shown, and differ from the steering angle of the motorcycle steering assembly and camber angle of the motorcycle body. Tire forces $F_{x,y,z}^t$ are discussed in Section \ref{sec:vehicle_tire_forces}.} 
    \label{fig:tire_diagram}
\end{figure}

So far, our geometric picture does not completely describe a motorcycle: we must additionally consider tire geometry. Unlike their four-wheeled counterparts, motorcycle tires must both camber and steer to large extents. Motorcycle motion is further complicated by the front steering assembly rake angle $\epsilon$ and offset $\delta$ (Figure \ref{fig:motorcycle_side_dims}), and the need to carefully define camber and steering of a tire. We follow the convention of \cite{pacejka_tire_book}, illustrated in Figure \ref{fig:tire_diagram}. The tire camber angle $c^t$ is the angle between the tire plane of symmetry and the body frame vertical $\eb{3}$, and the steering angle $\gamma^t$ is the angle between the $\eb{1}$ direction and the intersection of the tire plane of symmetry with the $\eb{1}-\eb{2}$ plane. These angles are not impacted by the concept of a camber axis, however they are coupled and dependent on steering head angle $\epsilon$, which we derive next. We use superscripts $^f$ and $^r$ in place of $^t$ for quantities specific to the front and rear tire respectively.

We assume the rear tire is unsteered, as a result $\gamma^r = 0$ and $c^r = c$; the rear tire cambers exactly with the body.

For the front tire, we must consider successive rotation first about the steering column with angle $\gamma$, itself rotated by angle $\epsilon$, and then the camber angle $c$ of the motorcycle body. To do so exactly we consider the wheel plane normal vector, $\et{2}$ in Figure \ref{fig:tire_diagram}. It relates to $c^t$ and $\gamma^t$ as:
\begin{subequations}
\begin{align}
    c^t &= -\sin^{-1}\paren{\et{2}\cdot \eb{3}} \\
    \gamma^t &= \tan^{-1}\paren{-\frac{\et{2} \cdot \eb{1}}{\et{2} \cdot \eb{2}}}.
\end{align}
\end{subequations}
For the front tire, these vector expressions follow from standard rotation theory and consideration of steering $\paren{\gamma}$, head angle $\paren{\epsilon}$, and camber $\paren{c}$:
\begin{subequations}
\begin{align}
        \ef{2} \cdot \eb{1} &= -\cos(\epsilon)\sin(\gamma) \\ 
        \ef{2} \cdot \eb{2} &=  \cos(c)\cos(\gamma) - \sin(c)\sin(\epsilon)\sin(\gamma)\\
        \ef{2} \cdot \eb{3} &= -\sin(c)\cos(\gamma) - \cos(c)\sin(\epsilon)\sin(\gamma).
\end{align}
\end{subequations}
Expressions for $c^f$ and $\gamma^f$ from $c$, $\gamma$, and $\epsilon$ are immediate:
\begin{subequations}
\begin{align}
    c^f &= \sin^{-1} \paren{\sin(c)\cos(\gamma) + \cos(c)\sin(\epsilon)\sin(\gamma)}\\
    \gamma^f &= \tan^{-1}\paren{\frac{\cos(\epsilon)\sin(\gamma)}{\cos(c)\cos(\gamma) - \sin(c)\sin(\epsilon)\sin(\gamma)}}.
\end{align}
\end{subequations}
 
The last tire quantity we consider explicitly is the slip angle, defined as the angle between the tire plane of symmetry and the motion of its point of contact with the road. 
This takes the form:
\begin{align} \label{eq:slip_angle} 
    \alpha^t &= \tan^{-1}\paren{\frac{\vt{2}\cos(\gamma^t) - \vt{1}\sin(\gamma^t) }{\vt{1}\cos(\gamma^t) + \vt{2} \sin(\gamma^t)}},
\end{align}
where $\bs{v}^t$ is found by translating velocity from the body frame origin to the tire contact point. For example the front tire velocity is related to the linear $\paren{\bs{v}^b}$ and angular $\paren{\bs{\omega}^b}$ velocity of the reference location as:
\begin{subequations}
\begin{align}
    \vf{1} &= \vb{1} - \wb{2} r \\
    \vf{2} &= \vb{2} + l_f \wb{3} + \wb{1} r.
\end{align}
\end{subequations}

Slip ratio and turn slip may be derived similarly. Slip ratio measures how far a tire is from spinning freely, while turn slip considers steering and yaw motion of a tire. Mathematical definitions of slip ratio and turn slip may be found in Section 2.2 of \cite{pacejka_tire_book}. We limit our attention to camber and slip angle effects as they are the dominant sources of lateral tire forces. Slip ratio is the dominant source of longitudinal tire forces, however we treat longitudinal tire force as an input in Section \ref{sec:vehicle_dynamics}.

\subsection{Road Surface Model}

\begin{figure}
    \centering
    \includegraphics[width=\linewidth]{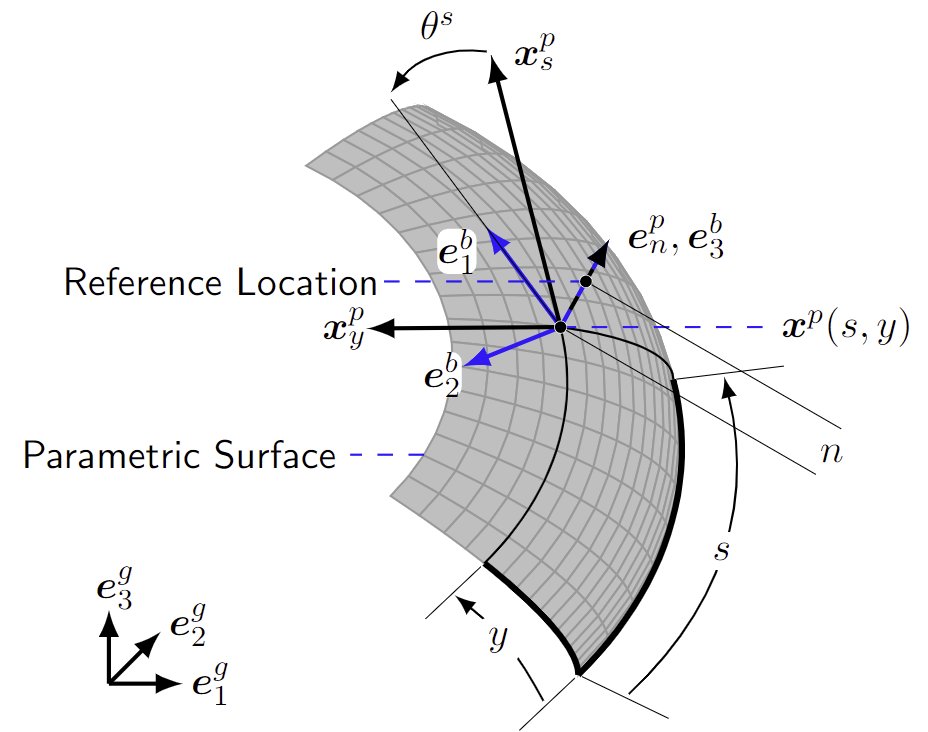}
    \caption{Schematic of parametric road surface model. The surface is defined by a function $\xp(s,y)$ and a vehicle reference location is located at a fixed normal offset $n$ from the surface. The reference location is equipped with the basis $\braces{\eb{1},\eb{2},\eb{3}}$, which remains tangent to the surface at all times ($\eb{3}=\xpn$). With these assumptions, position and orientation of the reference location and basis are fully determined by $\paren{s,y,\ths}$.}
    \label{fig:surface_defn}
\end{figure}

We leverage the 3D road surface model introduced in \cite{fork2021models} and extended in \cite{fork2023models}, with novel application to motorcycles. In \cite{fork2023models}, the body of a four-wheeled vehicle is assumed to remain tangent to the road surface at a fixed normal offset $n$ from it. The reference location is chosen as the vehicle's center of mass, with the basis $\eb{1,2,3}$ corresponding to the orientation of the body of the vehicle. Vehicle pose is described with variables ($s,y,\ths$) as illustrated in Figure \ref{fig:surface_defn}. 

However, motorcycles do not remain tangent to a road when they camber. Therefore, we choose a different reference location than \cite{fork2023models}: the reference location shown in Figure \ref{fig:motorcycle_side_dims}, along with the basis $\eb{1,2,3}$. We then use the road model of \cite{fork2023models} to describe the motion of the reference location over a general nonplanar surface. Additional variables $c$ and $d$ (Figure \ref{fig:motorcycle_dims}) are needed to describe the pose of the motorcycle and will appear in our vehicle model.

Our road model requires us to know the linear and angular velocity of the reference location and the basis $\eb{1,2,3}$. We will make these state variables of the motorcycle, with variables $\dot{c}$ and $\dot{d}$ added to describe the velocity of the whole motorcycle. Some velocity components will be simplified by our road model. Our choice of reference location will alter expressions for motorcycle dynamics, which we address in Section \ref{sec:vehicle_dynamics}.

Both road and motorcycle model implicitly assume that the curvature of the surface is sufficiently gradual relative to the length of the motorcycle. We remark that this is often a reasonable assumption on racetracks, but remains a limitation of our approach.

\section{Road Model} \label{sec:road_model}
We introduce the parametric surface $\xp(s,y)$ to describe the shape of a road surface, which is stationary with respect to an inertial frame. 
We assume the body frame remains tangent to $\xp$ at a normal offset $n$. Mathematically, we have:
\begin{align} \label{eq:tangent_contact_constraints}
    \xp + n \xpn  &= \bs{x} & 
    \eb{3} \cdot \xps &= 0 & 
    \eb{3} \cdot \xpy &= 0.
\end{align}
Here $\bs{x}$ is the position of the reference location (Figure \ref{fig:surface_defn}). $\xps$ and $\xpy$ are the tangent vectors of $\xp$: the partial derivatives with respect to $s$ and $y$, while $\xpn$ is the normal vector of the surface and is identical to $\eb{3}$. Furthermore, we define a notion of relative heading using the angle $\ths$:
\begin{align}\label{eq:parametric_angle}
    \cos(\ths) &= \frac{\eb{1}  \cdot \xps}{\xpsmag} &
    \sin(\ths) &= \frac{-\eb{2} \cdot \xps}{\xpsmag}.
\end{align}

These constraints allow $s,y,\ths$ to fully describe the 3D position and orientation of the reference location and the frame $\eb{1,2,3}$. However, they do not directly tell us how $s,y,\ths$ change with respect to time; from \cite{fork2023models} we recall the following results:
\begin{subequations} \label{eq:pose_kinematics}
\begin{align}
        \begin{bmatrix}
            \dot{s}\\
            \dot{y}
        \end{bmatrix}
        &=
    \begin{aligned}
        \left(\one - n \two\right)^{-1}
        \bfJ
        \begin{bmatrix}
            \vb{1} \\
            \vb{2}
        \end{bmatrix}
        & &~~~~~~
        \dot{n} &= \vb{3}
    \end{aligned}
    \\
    \dot{\theta}^s &=
      \wb{3} +
         \frac{\left(\xpss\times \xps\right)\cdot \xpn}{\xps \cdot \xps}\dot{s}
       + \frac{\left(\xpyy\times \xps\right)\cdot \xpn}{\xps \cdot \xps}\dot{y}.
\end{align}
\end{subequations}

Together, these relate linear and angular velocity of the body frame to how $s,y,\ths$ change over time. As in the planar case, the in-plane linear velocity components $\vb{1}$ and $\vb{2}$ and the normal angular velocity component $\wb{3}$ are most important. These will be state variables of the motorcycle. $\one$~and~$\two$ are the first and second fundamental forms \cite{differential_geometry_of_curves_and_surfaces} of the parametric surface $\xp$, while $\bfJ$ is a Jacobian between the parametric surface and the body frame, defined as:
\begin{equation}
    \bfJ = 
    \begin{bmatrix}
    \xps \cdot \eb{1} & \xps \cdot \eb{2}\\
    \xpy \cdot \eb{1} & \xpy \cdot \eb{2}
    \end{bmatrix}.
\end{equation}
$\bfJ$ may also be written as:
\begin{subequations}
\begin{align}
    \thp =& - \sin^{-1} \left( \frac{\xps \cdot \xpy}{\xpsmag~\xpymag} \right)\\
    \bfJ=& 
    \begin{bmatrix}
        \cos(\ths) \xpsmag & -\sin(\ths) \xpsmag\\
        \sin(\ths - \thp)\xpymag & \cos(\ths - \thp)\xpymag
    \end{bmatrix},
\end{align}
\end{subequations}
which is easier to compute from $s,y,\ths$.

Tangent contact constraints have several other implications. First, the vertical velocity component is zero since
\begin{equation} \nonumber
    \vb{3} = \dot{n} = 0.
\end{equation}
Second, the in-plane angular velocity of the body frame is related to linear velocity \cite{fork2023models}:
\begin{equation} \label{eq:angular_vel_constraint}
    \begin{bmatrix}
        -\wb{2} \\ \wb{1}
    \end{bmatrix}
    =\bfJ^{-1} \two~\left(\one - n \two\right)^{-1} \bfJ
    \begin{bmatrix}
    \vb{1} \\
    \vb{2}
    \end{bmatrix}.
\end{equation}
Finally, forces such as gravity require us to know $\eb{1,2,3}$ as a function of vehicle pose: $s,y,\ths$. From \cite{fork2023models} we have:
\begin{align} \label{eq:p2g_q}
    \begin{bmatrix}
        \eb{1} & \eb{2}
    \end{bmatrix}
    &= \begin{bmatrix}
        \xps & \xpy
    \end{bmatrix}
    \mathbf{I}^{-1} \mathbf{J}
    &
    \eb{3}&=\xpn.
\end{align}

Equation \eqref{eq:angular_vel_constraint} captures how the body frame must rotate to remain tangent to the parametric surface. It also results in centrifugal and Coriolis forces when applied to rigid body dynamics. For dynamics we also need expressions for $\dot{\omega}_1^b$ and $\dot{\omega}_2^b$. As in \cite{fork2023models} we make the approximation:
\begin{equation} \label{eq:angular_accel_constraint}
    \begin{bmatrix}
        -\dot{\omega}_2^b \\ \dot{\omega}_1^b
    \end{bmatrix}
    \approx \bfJ^{-1} \two~\left(\one - n \two\right)^{-1} \bfJ
    \begin{bmatrix}
    \dot{v}_1^b \\
    \dot{v}_2^b
    \end{bmatrix}.
\end{equation}
This approximation neglects the time-rate-of-change of the boldfaced quantities, which altogether capture the curvature of the surface. In effect, we have assumed that the curvature changes gradually.

This completes our 3D road model, the predominant change from \cite{fork2023models} being a difference in how the body frame is interpreted. Importantly, the body frame does not wholly descibe the motorcycle, which may camber within it. However, we can fully describe vehicle pose using $s,y,\ths, c, d$. Similarly, the vehicle's velocity is fully described by $\vb{1}, \vb{2}, \wb{3}, \dot{c}, \dot{d}$, albeit with complicated dynamics expressions for how these variables change. These expressions are derived next.

\section{Vehicle Model} \label{sec:vehicle_model}
We derive our vehicle model as follows:
\begin{enumerate}
    \item Compute the momentum, and time rate of change thereof, of the system as a function of model variables. 
    \item Use 1) and Newtonian mechanics to relate model variables to the net force and moment on the motorcycle.
    \item Equate net force and moment from mechanics to the force and moment on the vehicle obtained from tire forces, gravity, and drag.
\end{enumerate}
Our process fully captures the modeled motorcycle kinematics, dynamics, and forces and moments. However, it yields a differential algebraic equation of the form:
\begin{align}\label{eq:dae}
    \dot{z} &= f(z,u,a) &
    0       &= g(z,u,a).
\end{align}
This form emerges because $g$ captures the force and moment equalities of step 3) as an equality constraint. We have introduced $z$ as a set of differential states, $a$ a set of algebraic states, and $u$ a set of input quantities to the model. In the present case, $z$, $u$, and $a$ are:
\begin{subequations}
\begin{align} \label{eq:model_arguments}
    z &= \{s,y,\ths,\vb{1},\vb{2},\wb{3},c,\dot{c},d,\dot{d}\}\\
    a &= \{\dot{v}^b_1, \dot{v}^b_2, \dot{\omega}^b_3, \ddot{c}, F_z^f, F_z^r\}\\
    u &= \{\gamma, \ddot{d},F_x^f,F_x^r\}.
\end{align}
\end{subequations}
The terms $F_z^f$ and $F_z^r$ are the normal force on the front and rear tires, while $F_x^f$ and $F_x^r$ are the longitudinal force of the same tires, later treated as an input by the tire model used. 

Equation \eqref{eq:pose_kinematics} provides $\dot{s}$, $\dot{y}$ and $\dot{\ths}$ for $\dot{z}$ (equivalently $f$). The remaining elements of $\dot{z}$ are elements of $z$, $u$, or $a$.

We derive $g$ using the approach stated at the start of the section. We derive expressions for the net force and moment on the vehicle as a function of $z$, $u$, and $a$ as a result of Newtonian mechanics. We then derive a \textit{second} set of expressions for net force and moment as a result of gravity, tire forces, and aerodynamics. Then, $g$ equates the two: $g$ has three force equalities and three moment equalities. 

Implicitly, $g$ ensures that vehicle motion satisfies Newtonian mechanics subject to modeled sources of force and moment. In many specific cases one can solve for $a$ as a function of $z$ and $u$, and replace \eqref{eq:dae} with $\dot{z} = f(z, u, a(z,u))$. However, this is unnecessary from an implementation point of view\footnote{Broadly speaking, optimal control of \eqref{eq:dae} can be set up by solving an ordinary differential equation for $z$ with new variables for $a$ and equality constraints for $g$.} and poses numerical risks and challenges for finding $a(z,u)$ in closed form. Furthermore, the elements of \eqref{eq:dae} are sparse, and can be derived by a computer using symbolic algebra for the motorcycle model developed here, a useful aid as they are complicated.

\subsection{Vehicle Dynamics} \label{sec:vehicle_dynamics}

In this section we relate net force and moment to $z$, $u$, and $a$ using Newton's second law of motion:
\begin{align} \label{eq:newtons_second_law}
    \frac{d}{dt}\bs{p} &= \bs{F} &
    \frac{d}{dt}\bs{l} &= \bs{K},
\end{align}
where $\bs{p}$ and $\bs{l}$ are the linear and angular momentum of a system with external force $\bs{F}$ and moment $\bs{K}$. The terms $\bs{p}$, $\bs{l}$, and $\bs{K}$ are with respect to the center of mass. As our reference location on the vehicle is not the center of mass, these expressions are nontrivial and must be derived. We do so in a step-by-step manner, leaving the final result in compact form that depends on intermediate expressions. These steps can be expanded by hand or automated on a computer using symbolic algebra (the author's approach). 

\subsubsection{Net Force from Mechanics} \label{sec:mechanics}
To begin, the position of the COM relative to our reference location is:
\begin{equation}
    \rcom = \emot{3} (h-r) + \emot{2} d.
\end{equation}
The basis vectors $\emot{1,2,3}$ are in turn related to $\eb{1,2,3}$ as:
\begin{subequations}
\begin{align} \label{eq:motorcycle_frame_eqns}
    \emot{1} &= \eb{1} \\
    \emot{2} &= \eb{2} \cos\paren{c} - \eb{3} \sin\paren{c} \\
    \emot{3} &= \eb{2} \sin\paren{c} + \eb{3} \cos\paren{c},
\end{align}
\end{subequations}
and the time derivative of $\eb{1,2,3}$ with respect to the inertial frame are:
\begin{align}
    \begin{bmatrix}
        \frac{d}{dt}\eb{1} \\
        \frac{d}{dt}\eb{2} \\
        \frac{d}{dt}\eb{3}
    \end{bmatrix}
    =
    \begin{bmatrix}
        0 & \wb{3} & -\wb{2} \\
        -\wb{3} & 0 & \wb{1} \\
        \wb{2} & -\wb{1} & 0 \\
    \end{bmatrix}
    \begin{bmatrix}
        \eb{1} \\ \eb{2} \\ \eb{3}
    \end{bmatrix}.
\end{align}
Thus, the linear velocity of the center of mass is:
\begin{equation}
    \vcom = \bs{v}^b + \frac{d}{dt}\rcom ,
\end{equation}
where the rightmost term is found by chain rule:
\begin{equation}
\begin{split}
    \frac{d}{dt} \rcom = \dot{c} \partial_c \rcom  + \dot{d} \partial_d \rcom  \\+ \sum\nolimits_{k=1,2,3}\frac{d}{dt}(\eb{k}) \partial_{\eb{k}} \rcom .
\end{split}
\end{equation}

This procedure is tedious to evaluate by hand but can be automated by symbolic algebra with $\eb{k}$ treated as scalar symbols during the process. The net force on the vehicle, per Newton's second law, then follows from \eqref{eq:newtons_second_law}:
\begin{equation}
    \bs{F} = \frac{d}{dt}\bs{p} = m\frac{d}{dt} \vcom.
\end{equation}
Finding $\frac{d}{dt} \vcom$ involves one more round of differentiation with additional derivatives of $\vb{1}$, $\vb{2}$, $\wb{3}$, $\dot{c}$, and $\dot{d}$ in the chain rule\footnote{Equation \eqref{eq:angular_accel_constraint} and its assumptions are implicit here by ignoring the partial derivatives with respect to $s$ and $y$ of the boldfaced terms of \eqref{eq:angular_vel_constraint}, which are present as $\wb{1}$ and $\wb{2}$ factor in the COM velocity}. The result is a single expression with terms that include either $\eb{1}$, $\eb{2}$, or $\eb{3}$. The net force on the vehicle in direction $\eb{k}$ is then the sum of all terms that include $\eb{k}$.

\subsubsection{Net Moment from Mechanics}
For angular momentum we make the approximation that moment of inertia is constant, ie. $d$ is sufficiently small such the moment of inertia of the system does not change. However, the moment of inertia of the rider-vehicle system is only approximately constant in the motorcycle frame, not the body frame. As a result we first compute the angular velocity of the motorcycle frame with respect to the inertial frame:
\begin{subequations}
\begin{align}
    \wm{1} &= \wb{1} + \dot{c} \\ 
    \wm{2} &= \wb{2} \cos(c) - \wb{3} \sin(c) \\
    \wm{3} &= \wb{2} \sin(c) + \wb{3} \cos(c).
\end{align}
\end{subequations}
However, $\bs{\omega}^m$ is not the only source of angular momentum in a motorcycle: the engine and wheels contribute as well. We approximate the angular velocity ($\omega^t$) of either tire as the longitudinal velocity of the tire (see eqn. \ref{eq:slip_angle} and discussion of slip angle) divided by its radius, and approximate its angular momentum as:
\begin{equation}
    \bs{l}^t = \paren{\omega^t} I^t\emot{2} ,
\end{equation}
where we have assumed that the steering angle is small and $I^t$ is the rotational tire inertia, which may be increased for the rear wheel to approximate engine inertia as well. With this in mind, the angular momentum of the system can be approximated as:
\begin{equation}
    \bs{l} \approx
    \begin{bmatrix}
        \emot{1} \\ \emot{2} \\ \emot{3}
    \end{bmatrix}^T
    \begin{bmatrix}
        I^m_{11} & I^m_{12} & I^m_{13} \\ 
        I^m_{21} & I^m_{22} & I^m_{23} \\
        I^m_{31} & I^m_{32} & I^m_{33} \\
    \end{bmatrix}    
    \begin{bmatrix}
        \wm{1} \\ \wm{2} \\ \wm{3}
    \end{bmatrix}
     + \bs{l}^f + \bs{l}^r.
\end{equation}
Here we have introduced the the moment of inertia of the system in the motorcycle frame of reference: $I^m_{ij}$. The net moment in the body frame follows from Newton's second law \eqref{eq:newtons_second_law} and chain rule usage similar to the previous section. 

We remark that the procedure used here is quite general in nature and may be extended to other systems. For example, a single-wheeled robot or unicycle may be modeled by adding a second angle to consider pitch motion of the cycle. More complex driver motion may be considered by adding additional variables to describe their displacement or tilt. Any variables added must be considered during momentum computation, and added to $\eqref{eq:model_arguments}$ as appropriate.

\subsection{Vehicle Forces and Moments} \label{sec:vehicle_forces}

We discuss several common sources of force and moment on a motorcycle which we consider in this work:

\subsubsection{Gravity} \label{sec:vehicle_gravity_forces}
Gravitational forces on the motorcycle are fully determined by the 3D orientation of the body frame, which is determined by $s,y,\ths$ and is expressed in Eq. \eqref{eq:p2g_q}. Components of $\eb{1,2,3}$ in the direction of gravity ($\bs{g}$) then follow from \eqref{eq:p2g_q} and inner products with respect to $\bs{g}$:
\begin{subequations}
\begin{align}
    \begin{bmatrix}
        \eb{1} \cdot \bs{g} & \eb{2} \cdot \bs{g}
    \end{bmatrix}
    &= \begin{bmatrix}
        \xps \cdot \bs{g} & \xpy \cdot \bs{g}
    \end{bmatrix}
    \mathbf{I}^{-1} \mathbf{J}.
    \\
    \eb{3} \cdot \bs{g} &=\xpn \cdot \bs{g}.
\end{align}
\end{subequations}

\subsubsection{Aerodynamics} \label{sec:vehicle_drag}
Most aerodynamic models in control-oriented vehicle models relate current velocity to a force and moment. Two geometric considerations are needed here:
\begin{enumerate}
    \item The linear velocity $\bs{v}^b$ is close to the surface of the road. Rotational terms must be added when considering a different location on the motorcycle. 
    \item Forces and moments which act in direction $\emot{2}$ or $\emot{3}$ must be split into their $\eb{2,3}$ components as a function of camber angle (Eq. \eqref{eq:motorcycle_frame_eqns}).
\end{enumerate}

\subsubsection{Tire Forces} \label{sec:vehicle_tire_forces}

For the present case study we use the hypothetical motorcycle tire model proposed in \cite[ch. 11]{pacejka_tire_book}. This model treats the longitudinal force of a tire as an input, with peak lateral force diminished as a result. Furthermore, lateral tire force is a function of the normal force, camber angle, and slip angle of the tire. These quantities were defined and derived in Section \ref{sec:motorcycle_background} and can be obtained from variables in $z$, $u$, and $a$. Full equations and parameters may be found in Equations (11.40) through (11.59) and Table 11.1 of \cite{pacejka_tire_book}. We remark on several key features here. 

First, the tire forces $F_{x,y,z}^t$ are relative to the road (Figure \ref{fig:tire_diagram}). For instance the normal force is normal to the road and thus produces a moment about the center of mass when the motorcycle cambers. Second, the peak lateral force is modified by the longitudinal tire force $F_x^t$ (treated as an input), normal force $F_z^t$, camber angle $c^t$, and tire model parameters $d_4$ and $d_7$:
\begin{align}
    F_{y,\text{max}}^t &= \sqrt{D_0^2 - \paren{F_x^t}^2} &
   D_0 &= \frac{d_4 F_z^t}{1+d_7(c^t)^2}.
\end{align}

Body-frame components of tire forces follow from their steering angle (Section \ref{sec:motorcycle_background}). Moments produced by tire forces about the center of mass follow from their position relative to the COM. These vary with $c$ and $d$, which vary over time. For example, the position of the front tire contact patch relative to the COM is 
\begin{equation}
    l_f \eb{1} -r \eb{3} - (h-r) \emot{3} - d \emot{2},
\end{equation}
from which front tire moments may be computed.

As a result, we can compute expressions for the net force and moment as a result of tires, aerodynamics, and gravity. We obtain $g$ in \eqref{eq:dae} by equating these to the force and moment expressions that resulted from mechanics in Section \ref{sec:mechanics}, which completes our vehicle model. Other sources of force and moment may be seamlessly considered by adding additional equations which model their effects.

\begin{figure*}
    \centering
    \includegraphics[width=0.8\linewidth]{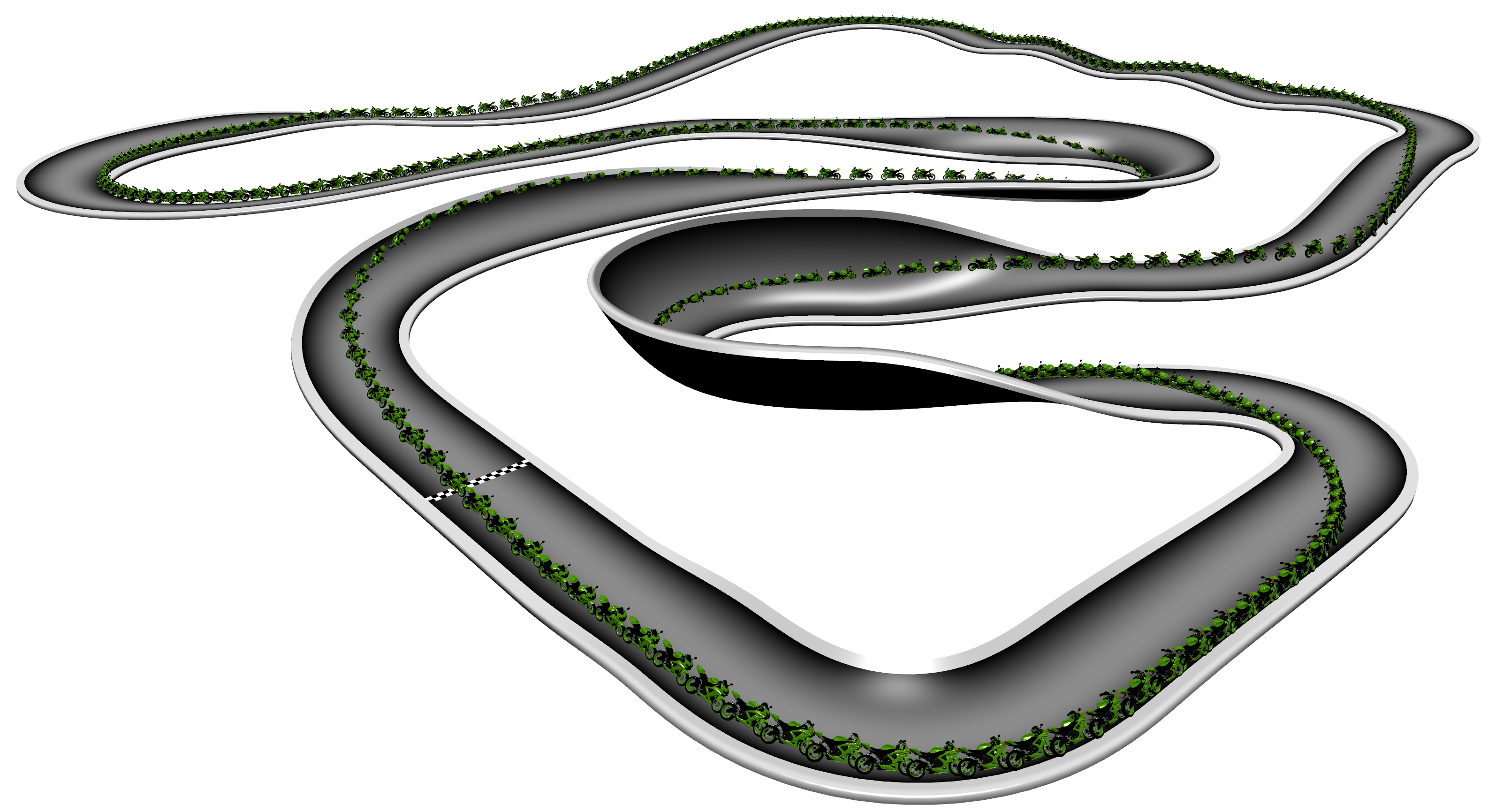}
    \caption{Nonplanar racetrack with snapshots of the motorcycle raceline shown every tenth of a second.}
    \label{fig:racetrack}
\end{figure*}

\section{Application to Optimal Control}  \label{sec:control}

We applied our nonplanar motorcycle model to compute racelines: periodic minimum time trajectories around a racetrack subject to the dynamics of the motorcycle model. We used direct collocation with Legendre polynomials, discussions of which can be found in \cite{biegler_book} or \cite{gpops_ii}. We fixed collocation intervals at uniform intervals of the $s$ coordinate along a periodic parametric surface $\xp$ with continuity constraints between the start and the end of the raceline. This matches earlier vehicle raceline literature \cite{3d_part_2, LOT20147559, leonelli2020optimal} with the key difference being the use of our novel vehicle model which allowed us to consider more general racetracks. We set up the surface, motorcycle model, and raceline problem in CasADi \cite{casadi}. We solved the raceline optimization problem with the nonlinear optimization solver IPOPT \cite{ipopt} and the linear solver MUMPS \cite{mumps}. 

To facilitate optimization we normalized decision variables for $F_z^f$, $F_z^r$, $F_x^f$, and $F_x^r$ in \eqref{eq:model_arguments} by the weight of the vehicle, and unscaled these variables when used as forces, such as in $g$ in \eqref{eq:dae}. Furthermore, we limited the input longitudinal force magnitude on each tire by the associated tire normal force. We also constrained both tire normal forces to be non-negative, which avoided cases where loss of road contact would occur. Finally, we approximated the power required to drive the rear wheel as $F_x^r \vb{1}$, which we upper bounded by parameter $P_\text{max}$. Model parameters are reported in Table \ref{tab:parameter-table}.

\begin{table}
\caption{Motorcycle Model Parameters}
\begin{tabular}{c c c c c c }
\hline
$m$ & 240 \unit{\kilo\gram}                          & $l_f$ & 0.75 \unit{\meter} & $ \gamma$ $\in\pm$ 0.7 & \unit{\radian}\\
$I^m_{11}$ & 18 \unit{\kilo\gram\square\meter}       & $l_r$ & 0.75 \unit{\meter} & $ \ddot{d}$ $\in\pm$ 0.5 &\unit{\meter\per\second\squared}\\
$I^m_{22}$ & 60 \unit{\kilo\gram\square\meter}       & $h$ & 0.5 \unit{\meter} & $ {d}$ $\in\pm$ 5 &\unit{\centi\meter}\\
$I^m_{33}$ & 48 \unit{\kilo\gram\square\meter}       & $r$ & 0.1 \unit{\meter} & $P_\text{max}$& 50 \unit{\kilo\watt}\\
$\abs{\bs{g}}$ & 9.81 \unit{\kilo\gram\per\square\second}       & $\epsilon$ & $30^{\circ}$\\
\hline
\end{tabular}
\label{tab:parameter-table}
\end{table}

\section{Results}  \label{sec:results}

We computed a raceline on a hypothetical racetrack which could not be studied with earlier methods due to its curved cross-section. Racetrack and raceline are shown in Figure \ref{fig:racetrack}. The $650$ meter long track includes many nonplanar features such as quarter-pipe turns, gullies, and undulating hills. Our method achieved a lap time of $31.1$ seconds, with $33$ seconds of compute time to converge to local optimality on an 11\textsuperscript{th} Gen Intel\textregistered\ Core\textsuperscript{TM} i7-11800H @2.3GHz. Although optimization can only guarantee a locally-optimal trajectory, intuitive 3D behaviour of the motorcycle is emergent from our model. For example, planar turns are all made on the inside of the turn, while nonplanar turns appear on the inside, outside or middle of the turn, depending on the geometry of the racetrack.

\section{Conclusion} \label{sec:conclusion}
We extended the interpretation of our general nonplanar road model to apply it to the dynamics of motorcycles. In the process we added considerations for motorcycle camber and rider motion, and their impact on motorcycle dynamics. We discussed how the symbolic complexity of the vehicle model can be addressed by computer algebra. We used our model to generate time-optimal racelines on a complex nonplanar racetrack, and illustrated that intuitive racing behaviour emerged from use of our nonplanar motorcycle model.

\section{Acknowledgements}
 The visual \href{https://skfb.ly/oIYFv}{\textcolor{blue}{\underline{\smash{3D motorcycle model}}}} in Figure \ref{fig:racetrack} was made by Sketchfab user nouter2077 and is licensed \href{https://creativecommons.org/licenses/by/4.0/}{{\textcolor{blue}{\underline{\smash{CC BY 4.0}}}}}.

\clearpage
\printbibliography

@article{leonelli2020optimal,
  title={Optimal control of a road racing motorcycle on a three-dimensional closed track},
  author={Leonelli, L and Limebeer, DJN},
  journal={Vehicle system dynamics},
  volume={58},
  number={8},
  pages={1285--1309},
  year={2020},
  publisher={Taylor \& Francis}
}

@incollection{tanelli2014modelling,
  title={Motorcycle Dynamics},
  author={Tanelli, Mara and Corno, Matteo and Saveresi, Sergio},
  year={2014},
  publisher={John Wiley \& Sons},
  booktitle   = {Modelling, Simulation and Control of Two-Wheeled Vehicles},
  pages       = "3-42",
  chapter     = 1,
}

@article{LOT20147559,
title = {A Curvilinear Abscissa Approach for the Lap Time Optimization of Racing Vehicles},
journal = {IFAC Proceedings Volumes},
volume = {47},
number = {3},
pages = {7559-7565},
year = {2014},
note = {19th IFAC World Congress},
issn = {1474-6670},
author = {R. Lot and F. Biral}
}

@article{3d_part_1,
    author = {Perantoni, Giacomo and Limebeer, David J. N.},
    title = "{Optimal Control of a Formula One Car on a Three-Dimensional Track—Part 1: Track Modeling and Identification}",
    journal = {Journal of Dynamic Systems, Measurement, and Control},
    volume = {137},
    number = {5},
    year = {2015},
    month = may
}

@article{3d_part_2,
    author = {Limebeer, D. J. N. and Perantoni, G.},
    title = "{Optimal Control of a Formula One Car on a Three-Dimensional Track—Part 2: Optimal Control}",
    journal = {Journal of Dynamic Systems, Measurement, and Control},
    volume = {137},
    number = {5},
    year = {2015},
    month = may}

@article{limebeer2023review,
  title={A review of road models for vehicular control},
  author={Limebeer, DJN and Warren, E},
  journal={Vehicle system dynamics},
  volume={61},
  number={6},
  pages={1449--1475},
  year={2023},
  publisher={Taylor \& Francis}
}

@INPROCEEDINGS{7225830,
    author={J. {Kong} and M. {Pfeiffer} and G. {Schildbach} and F. {Borrelli}},
    booktitle={2015 IEEE Intelligent Vehicles Symposium (IV)}, 
    title={Kinematic and dynamic vehicle models for autonomous driving control design}, 
    year={2015},
    volume={},
    number={},
    pages={1094-1099}}

@ARTICLE{7490340,
    author={B. {Paden} and M. {Čáp} and S. Z. {Yong} and D. {Yershov} and E. {Frazzoli}},
    journal={IEEE Transactions on Intelligent Vehicles}, 
    title={A Survey of Motion Planning and Control Techniques for Self-Driving Urban Vehicles}, 
    year={2016},
    volume={1},
    number={1},
    pages={33-55}}

@article{christ2021time,
  title={Time-optimal trajectory planning for a race car considering variable tyre-road friction coefficients},
  author={Christ, Fabian and Wischnewski, Alexander and Heilmeier, Alexander and Lohmann, Boris},
  journal={Vehicle system dynamics},
  volume={59},
  number={4},
  pages={588--612},
  year={2021},
  publisher={Taylor \& Francis}
}

@book{rajamani_book,
    author={Rajesh Rajamani},
    title={Vehicle Dynamics and Control}, 
    subtitle = {},
    publisher = {Springer},
    year={2011},
    edition={2nd}}

@book{guiggiani_book,
    author = {Massimo Guiggiani},
    title = {The Science of Vehicle Dynamics},
    subtitle = {Handling, Braking, and Ride of Road and Race Cars},
    publisher = {Springer},
    address = {Cham, Switzerland},
    year = {2023},
    edition = {3rd}
    }

@book{reza_book,
    author = {Reza N. Jazar},
    title = {Vehicle Dynamics},
    subtitle = {Theory and Application},
    publisher = {Springer},
    year = {2007},
}

@book{differential_geometry_of_curves_and_surfaces,
    author = {Manfredo P. do Carmen},
    title = {Differential geometry of curves and surfaces},
    publisher = {Prentice Hall},
    year = {1976}
}

@inproceedings{fork2021models,
      title={Models and Predictive Control for Nonplanar Vehicle Navigation}, 
      author={Thomas Fork and H. Eric Tseng and Francesco Borrelli},
      year={2021},
      booktitle = {2021 IEEE International Intelligent Transportation Systems Conference (ITSC)},
      pages = {749-754}
}

@article{fork2023models,
  title={Models for ground vehicle control on nonplanar surfaces},
  author={Fork, Thomas and Tseng, H Eric and Borrelli, Francesco},
  journal={Vehicle System Dynamics},
  pages={1--25},
  year={2023},
  publisher={Taylor \& Francis}
}

@book{pacejka_tire_book,
    author = {Hans Pacejka},
    title = {Tire and Vehicle Dynamics},
    edition = {3rd},
    year = {2012},
    publisher = {Butterworth-Heinemann}}

@book{biegler_book,
    author = {Lorentz T. Biegler},
    title = {Nonlinear Programming},
    subtitle = {Concepts, Algorithms, and Applications to Chemical Processes},
    year = {2010},
    publisher = {SIAM}}

@article{gpops_ii,
author = {Patterson, Michael A. and Rao, Anil V.},
title = {GPOPS-II: A MATLAB Software for Solving Multiple-Phase Optimal Control Problems Using Hp-Adaptive Gaussian Quadrature Collocation Methods and Sparse Nonlinear Programming},
year = {2014},
issue_date = {October 2014},
publisher = {Association for Computing Machinery},
address = {New York, NY, USA},
volume = {41},
number = {1},
journal = {ACM Trans. Math. Softw.},
month = {10},
articleno = {1},
numpages = {37}
}

@Article{casadi,
  Author = {Joel A E Andersson and Joris Gillis and Greg Horn
            and James B Rawlings and Moritz Diehl},
  Title = {{CasADi} -- {A} software framework for nonlinear optimization
           and optimal control},
  Journal = {Mathematical Programming Computation},
  Year = {2018},
}

@article{ipopt,
author = {Wächter, Andreas and Biegler, Lorenz},
year = {2006},
month = {03},
pages = {25-57},
title = {On the Implementation of an Interior-Point Filter Line-Search Algorithm for Large-Scale Nonlinear Programming},
volume = {106},
journal = {Mathematical programming}
}

@article{mumps,
   title   = {A Fully Asynchronous Multifrontal Solver Using Distributed Dynamic Scheduling},
   author  = {P.R. Amestoy and I. S. Duff and J. Koster and J.-Y. L'Excellent},
   journal = {SIAM Journal on Matrix Analysis and Applications},
   volume  = {23},
   number  = {1},
   year    = {2001},
   pages   = {15-41}
 }
\end{document}